\documentclass[conference]{IEEEtran}
\IEEEoverridecommandlockouts
\usepackage{cite}
\usepackage{amsmath,amssymb,amsfonts}
\usepackage{algorithmic}
\usepackage{graphicx}
\usepackage{textcomp}
\usepackage{xcolor}

\usepackage{url}

\def\BibTeX{{\rm B\kern-.05em{\sc i\kern-.025em b}\kern-.08em
    T\kern-.1667em\lower.7ex\hbox{E}\kern-.125emX}}
\begin{document}

\title{A Sketch-Based Neural Model for Generating Commit Messages from Diffs\\
}

\author{\IEEEauthorblockN{Nicolae-Teodor Pavel}
\IEEEauthorblockA{University Politehnica of Bucharest \\
Romania \\
nicolae.teodor.pavel@gmail.com}
\and
\IEEEauthorblockN{Traian Rebedea}
\IEEEauthorblockA{University Politehnica of Bucharest \\
Romania \\
traian.rebedea@cs.pub.ro }
}


\maketitle

\begin{abstract}
Commit messages have an important impact in software development, especially when working in large teams. Multiple developers who have a different style of writing may often be involved in the same project. For this reason, it may be difficult to maintain a strict pattern of writing informative commit messages, with the most frequent issue being that these messages are not descriptive enough. In this paper we apply neural machine translation (NMT) techniques to convert code diffs into commit messages and we present an improved sketch-based encoder for this task. We split the approach into three parts. Firstly, we focus on finding a more suitable NMT baseline for this problem. Secondly, we show that the performance of the NMT models can be improved by training on examples containing a specific file type. Lastly, we introduce a novel sketch-based neural model inspired by recent approaches used for code generation and we show that the sketch-based encoder significantly outperforms existing state of the art solutions. The results highlight that this improvement is relevant especially for Java source code files, by examining two different datasets introduced in recent years for this task.
\end{abstract}

\begin{IEEEkeywords}
commit message generation, code summarization, neural machine translation, sketch-based encoder, deep learning
\end{IEEEkeywords}

\section{Introduction}

Version control is widely used in software development to maintain the 
history of changes for most software projects nowadays. Version control systems have been adopted by a large spectrum of developers, ranging from industry to academia. Each change is called a commit and should be described using a commit message. The purpose of this message is to offer a short
description of the changes. Commit messages are a key factor in software development, especially when working in teams, because they help different developers to understand, review and validate the commits. However, every developer has its own way of writing
the commit messages and it is difficult to maintain a strict pattern for writing informative short texts. Consequently, these messages are sometimes too short, too cryptic, or may be even left empty by some developers.

A solution to this problem is to automatically generate messages by examining the underlying source code diffs from each commit. Most of the systems \cite{Documenting, Towards, ChangeScribe}
that were previously developed for automatically generating commit messages do not use neural networks but rather more 
conventional ways, including rules and patterns, to create the messages. These systems proved to have poor performance and multiple limitations. For example Jiang et al. \cite{Towards} proposed a Naive Bayes approach for generating the verb of a commit message. Their approach was able to achieve only a 43\% accuracy. 

In this paper, we improve the work of Jiang et al. \cite{Automatically} who proposed the usage of neural machine translation (NMT) models to
"translate" code diffs to commit messages. Diffs are files that track the changes between two versions of a software repository. Besides the actual code changes, diff files contain some additional context about those changes as well. 

In our
experiments we used two large datasets containing diffs and commit messages  released in recent year. The former was proposed by Jiang et al. \cite{Towards} and the latter was released by Liu et al. \cite{How} which is a cleaned subset of the first one. Our most important aim is to determine the full potential of the recent advances in neural models for code generation and machine translation when applied for commit message generation.

To this extent, our study focuses on answering the following three research questions.

\textbf{RQ1: Has NMT achieved its full potential for commit message generation?}

We firstly focused on finding a more suitable NMT baseline for generating commit messages from diff files by extending the work of Jiang et al. \cite{Automatically}. For this, we explored three
verticals that can increase the performance of NMT models: model depth, embedding dimensionality, and residual connections.  
In addition to their work, we also tackle the problem of rare words by using BPE\cite{BPE} preprocessing and coping mechanism \cite{Copying}. By approaching this problem we aim at making our solution capable of open vocabulary translation from diffs to messages. 

\textbf{RQ2: Can training different NMT models for specific file types improve performance?}

We discovered a NMT model which is capable of obtaining state of the art performance. However, this NMT model is still not capable to outperform NNGen, a solution based on nearest neighbor search for commit messages \cite{How}. To understand why NMT models are outperformed by NNGen, we conducted a more thorough analysis of the diff files in the dataset. Thus, we discovered that the diff files contain a mix of file types extracted from code repositories and each diff contains only the changes of a single file. We decide to select the top nine most frequent file types and create a separate dataset for each file type. The training process on these datasets is similar to training on the entire dataset. The ensemble of NMT models trained on these specific file type datasets outperforms NNGen on the initial dataset but not on the cleaned one. Our experimental results show that models with fewer layers are able to generate state of the art results. 

\textbf{RQ3: Can a sketch-based neural model outperform NNGen for generating commit messages for Java files?}

During the previous experiments, we conducted a comparison by file type with NNGen which revealed that it outperforms NMT models by a high margin on examples containing Java files. Inspired by this finding, we propose a novel sketch-based neural model which outperforms NNGen for examples containing Java files. The preprocessing part for the proposed 
sketch-based approach generates a simple code sketch by replacing constants, variables, functions, and class names with placeholders. The
training is done similar to the previous NMT experiments using the code sketch. After training, the placeholders are replaced  with
constants, variables, functions, and class names. The ensemble containing the model trained with the sketch-based approach for Java files and the previous NMT models trained on each file type is able to outperform NNGen on both datasets.


In the rest of this paper, we describe the related work, the proposed method focusing on the sketch-based model, the experimental setup, and a summary of the results. At the end, we present both a quantitative comparison with the state of the art, together with a qualitative study using human evaluation and  concluding remarks on using advanced NMT models for commit message generation.

\section{Related work}\label{RL}

\subsection{NMT Models for Commit Message Generation}

Jiang et al. \cite{Automatically} proposed a neural machine translation (NMT) model to generate commit messages from diffs, an approach previously used in natural language translation. They worked with the dataset published by Jiang and McMillan \cite{Towards} consisting of diffs and their respective commit messages from several Java repositories collected from GitHub. The source code diffs are considered as the source language and the commit messages, filtered by keeping an action verb and direct object structure, as target language.

Their model consists of two layers, one for encoder and the other for decoder, both being recurrent neural networks. They employ a standard Bahdanau attention mechanism \cite{Attention}. The word embeddings size used by them is 512 and the hidden layers size is 1024. They train the model using early stopping for maximum 5000 epochs with a batch size of 80.

They obtain a BLEU-4 \cite{BLEU} score of 31.92. Compared to this approach, we are proposing an improved neural architecture using a sketch-based encoder, but also a more complex approach which divides the dataset into smaller datasets based on the file type found in the diff file and training an ensemble of models for each file type specific dataset.

\subsection{NNGen - Nearest Neighbour Commit Message Generation}

Liu et al. \cite{How} have analyzed in more depth how the neural machine translation approach proposed by
Jiang et al. \cite{Automatically} performs. They have analyzed when and why
a NMT model performs well and proposed a simpler solution inspired by the
nearest neighbour algorithm.

Firstly, they discovered that the approach of Jiang et al. generates very good messages on diffs
that have a high similarity with diffs in the training set. They also discovered that 16\%
of the messages were automatically generated or describe changes that are trivial. 

Taking into consideration these findings, Liu et al. created a new dataset by cleaning 
the dataset proposed by Jiang et al. \cite{Towards}. Their cleaning is done by manually evaluating 200 randomly picked commits from Jiang et al. test set. They determine that 71 (35.5\%) of those messages have good quality. Out of those 71 messages, all messages except one have high similarity with messages from the training set.

Lastly, Liu et al. put forward a new method called NNGen based on the nearest neighbour (NN) algorithm. 
Their approach outperforms Jiang et al. \cite{Automatically} approach 
on both datasets. Their approach obtains 38.55 (vs. 31.92) BLEU-4 score on the initial dataset
and 16.42 (vs. 14.19) BLEU-4 score on the cleaned dataset.

In our work we will first show that Liu et al. compared their results with a week NMT baseline. We will then describe how to build a NMT approach that has a similar performance as NNGen and how to subsequently improve it using a sketch-based NMT model.

\subsection{Neural Machine Translation}

Our work is mostly related to approaches in natural language processing that address the problems using neural machine translation.  We took an in-depth look at articles that explore the current state of the art for neural machine translation \cite{Massive, Google}. Currently the most used architecture is the sequence-to-sequence model \cite{seq2seq, Massive, Google, Copying}. It consists of two recurrent neural networks, an encoder and a decoder. The encoder and decoder can have one or more layers. Residual connections are commonly used to improve the performance of models with more than one layer for decoder or encoder.

The encoder-decoder approach needs to encode a sentence to a fixed size vector space. This can be a difficult task for sentences that are too long. In order to address this issue, the model is extended. Attention mechanisms \cite{Attention} are added to the model in order to learn to align the input sequence with the output sequence. Bahdanau\cite{Attention} attention is one of the most used attention mechanisms.

NMT models should be able to be robust and to translate words that are out of the vocabulary. There are two common techniques that address this issue: copying mechanism \cite{Copying} and subword units \cite{BPE}. Copying mechanism is learning to output out of vocabulary (OOV) words by copying them directly from the input. The model learns when to generate words from the existing vocabulary and when to copy words from the input. The subword units approach has the main goal of dividing rare words into subunits which occur more frequently. By doing so, the model generates rare words or OOV words from subunits learned during training.

\subsection{Sketch-based Methods for Code Generation}

Sketch-based methods have been recently found to substantially increase the performance of vanilla NMT models for code generation \cite{sketch1, sketch2}. Both methods are using sequence-to-sequence models, also employed in NMT, to generate sketches for source code \cite{sketch1} and SQL queries \cite{sketch1, sketch2}. Afterwards a different model is used in order to generate the final result (code or query). For example, Hosu et al. \cite{sketch2} use another sequence-to-sequence model with a dual encoder which receives both the sketch and the question as input in order to generate the final SQL query. In contrast to their approaches we leverage the coding style to create a rule-based method that creates the sketches. The decoding part in our case is performed using a dictionary created during the sketch generation. Moreover, in our case we have a sketch-based encoder, while these methods employ a sketch-based decoder for the sequence-to-sequence model.

\section{Proposed Method}\label{sec:proposed-method}

Our proposed method is split in three parts. Firstly, we focus on finding a more suitable NMT baseline. Secondly, we propose an improved method which relies on training on smaller datasets which contain the same file type. Lastly, we introduce a novel sketch-based neural encoder which substantially improves the performance of standard NMT models when generating commit messages for Java files. All our code is available on GitHub and can be found at the following link \footnote{\url{https://github.com/nicolae-teodor-pavel/commit-message-generation}}.

\subsection{Baseline NMT Method}
In this section we focus at first on NMT architecture exploration for our task.
Afterwards we tackle the problem of  rare and out of the vocabulary words. 

\paragraph{Architecture exploration}
One of our main goals is to find an architecture that better suits this problem 
which is very different from the problem of translating from one 
language to another. We explore how different variables
of the model affect the overall performance. The problem of translating
diffs to language is more special than (natural) language to (natural) language translation, 
therefore the need for advanced models is justified. We explore how the depth of 
the model influences the performance by testing with 2, 4 and 8 layers. Diffs are a 
combination of comments and code and our intuition says that a bigger embedding 
will help the model. In that regard we conduct tests with an embedding of size 512 
and 1024. In recent years, residual connections have helped deep models to obtain 
better performances. Although our model has less layers than Resnet \cite{ResNet} 
which introduced residual connections, we also explore if residual the gradient flow helps 
our models to gain better performances.

\paragraph{Rare and OOV words}\label{par:rareoov}
BPE is a preprocessing method which is making a vocabulary compression
based on the frequency of the n-grams on the sources and targets texts.
During our experiments we are going to test a vocabulary size of 5,000,
10,000 and 32,000. Based on the vocabulary size, the input length is 
different. We selected the input sequence length in order to contain 99\% of the 
examples in the dataset. The output sequence length remains constant at 30. In Table \ref{tab:bpe-versions} we present the three BPE versions used in our experiments.

\begin{table}[htbp]
\caption{BPE versions for processing rare and OOV words}
\begin{center}
\begin{tabular}{|l|c c c c|}
\hline
 & Vocabulary & Input sequence & Enc & Dec  \\
\hline
BPE1 & 5000 & 185  & 2 & 2 \\
BPE2 & 1000 & 170  & 4 & 4 \\
BPE3 & 32000 & 160  & 4 & 4\\
\hline
\end{tabular}
\end{center}

\label{tab:bpe-versions}
\end{table}

The copying mechanism implies reproducing parts from the input
sequence in the output sequence, as a way to replicate the human
way of speaking. After taking a closer look at the dataset we 
found that the initial dataset does not contain any out of the 
vocabulary word in the training messages. In order to use the 
copying mechanism, we reduced the messages and diffs
vocabularies based on the number of occurrences of each word.
In other words, we selected the most frequent words from both 
vocabularies. We tested two different configurations:


\begin{enumerate}
  \item The messages vocabulary is reduced to 6,302 which is 
  38.67\% from the initial messages vocabulary. The diffs
  vocabulary is not reduced.
  \item The messages vocabulary is reduced to 1,600 which is 
  9.81\% from the initial messages vocabulary and the diffs 
  vocabulary is reduced to 5,000 which is 
  10\% from the initial messages vocabulary.
\end{enumerate}

In the first configuration the messages vocabulary contains all the 
words that appeared in messages at least once in the train 
messages. In the second configuration the messages vocabulary 
contains the words that appeared at least 10 times in the train
messages. In the last configuration the diffs vocabulary 
contains the words that appeared at least in 10 times in the train diffs.


\subsection{Improved Ensemble NMT Model}

The next step in our work was to take a closer look on the dataset introduced by Jiang et al. \cite{Towards}.
One of the first observations was that all the diffs in the dataset contain only a change to a single file.
Following this observation we explored the frequency of each file type. We observed a 
mix of files in the dataset. In Table \ref{tab:file-type-freq} we present the top 10 most frequent file types in the original dataset \cite{Towards} and in the cleaned dataset \cite{How}. 80.49\% of the examples removed from the latter contain changes to Gitrepo files which are usually very simple to learn and not very relevant for source code commit messages.

\begin{table}[htbp]
\caption{Top 10 file types sorted by frequency}
\begin{center}
\begin{tabular}{|l| c c| c c|}
\hline
Dataset & \multicolumn{2}{c|}{Original \cite{Towards}} & \multicolumn{2}{c|}{Cleaned \cite{How}} \\
\hline
Rank & File type & Frequency & File type & Frequency  \\
\hline
1 & Java & 4186 & Java & 4186 \\
2 & Gitrepo & 3297 & Xml & 2097 \\
3 & Xml & 2103 & Gradle & 1817 \\
4 & Gradle & 1945 & Md & 1193 \\
5 & Md & 1619 & Gitignore & 912 \\
6 & Gitignore & 940 & Properties & 869 \\
7 & Properties & 899 & Txt & 820 \\
8 & Txt & 845 & Yml & 806 \\
9 & Yml & 806 & Js & 337 \\
10 & Js & 337 & Groovy & 297\\
\hline
\end{tabular}
\end{center}
\label{tab:file-type-freq}
\end{table}

Taking into account these observations, we explored how training a NMT model on a specific 
file type influences its performance. In our exploration we evaluated two scenarios. 
In the first one we select the top five most frequent
file types and create a dataset for each file type while in the second one we select 
the top nine most frequent file types. In both scenarios we group the rest of the 
files into a dataset which we will call "others".

The Gitrepo, Gradle, Java, Md and Xml represent 50.17\% of the examples in the dataset.
The Gitignore, Gitrepo, Gradle, Java, Md, Properties, Txt, Xml and Yml represent 63.48\% 
of the examples in the dataset. The vocabulary for each dataset was obtained following these steps:

\begin{enumerate}
    \item the words which do not appear in the initial vocabulary are filtered;
    \item the words which appear only once (twice for Gitrepo) in the diffs/messages specific 
    to the file types are filtered;
    \item the remaining words are selected to be part of the vocabulary.
\end{enumerate}

\begin{figure*}[htbp]
\centering
\begin{minipage}{.48\textwidth}
  \centering
  \includegraphics[width=\textwidth]{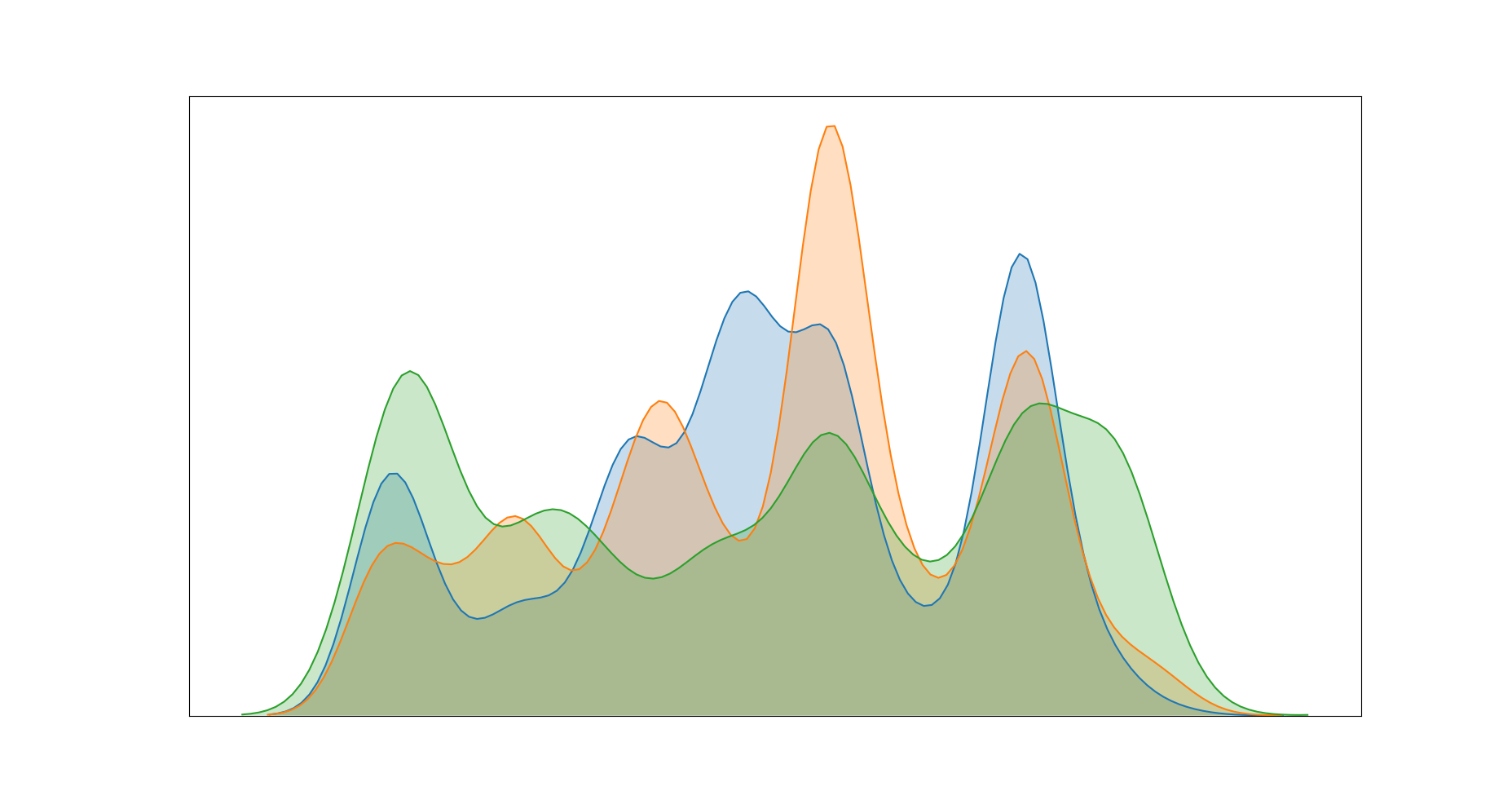}
  \caption{Kernel Density Estimation for Gitrepo, Java, Xml diff files; Gitrepo KDE is BLEU; 
  Java KDE is orange; Xml KDE is green}
  \label{fig:op-kde-diffs}
\end{minipage}%
 \hfill 
\begin{minipage}{.48\textwidth}
  \centering
  \includegraphics[width=\textwidth]{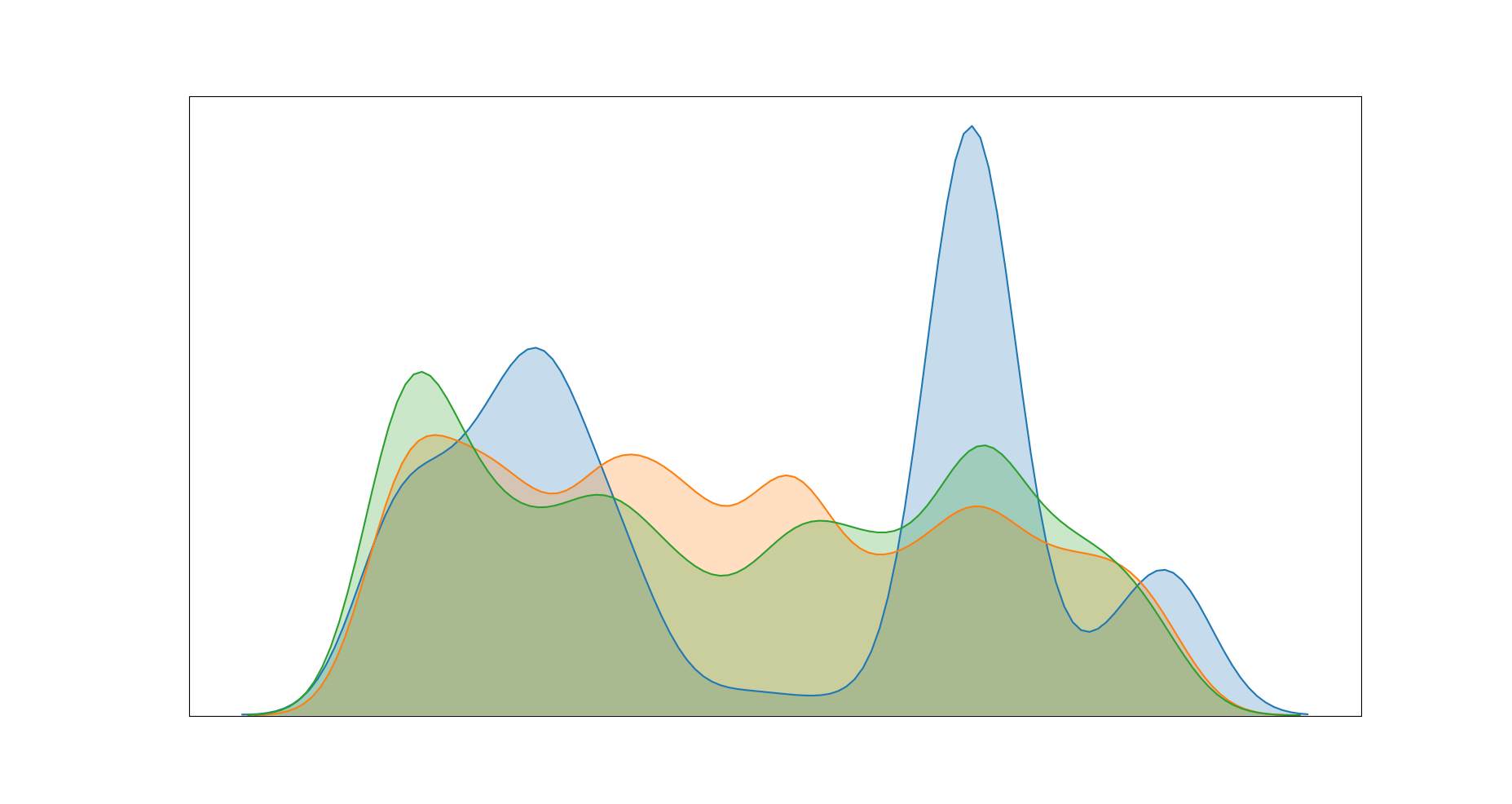}
  \caption{Kernel Density Estimation for Gitrepo, Java, Xml messages files; Gitrepo KDE is BLEU; 
  Java KDE is orange; Xml KDE is green}
  \label{fig:op-kde-msgs}
\end{minipage}
\end{figure*}

In order to confirm that our approach is valid, we used kernel density estimation (KDE) 
\cite{Kernel} to plot the density of words for the top
three most frequent file types (Gitrepo, Java, Xml). To obtain a relevant comparison we 
selected the first 10,000 words from the diffs and messages. Figure \ref{fig:op-kde-diffs} 
and Figure \ref{fig:op-kde-msgs} show that the distribution of the words in  diffs and messages is different for each file type.

\subsection{Sketch-based Neural Encoder}\label{subsec:sketc-proprosed}
Inspired by the finding of the poor performances of NMT on examples containing Java files,
we propose our main contribution which is a sketch-based neural model which achieves a better BLEU-4 score than
NNGen. One of the reasons we focus on examples containing Java files is that this file 
type is the only one which contains actual source code. Another reason is that the dataset
introduced by Jiang el al. was created from the most popular Java repositories on GitHub and any claim of
improved performance for this dataset should also report the score on examples containing Java file types. 

\begin{figure}[htbp]
  \begin{center}
  \includegraphics[width=\linewidth]{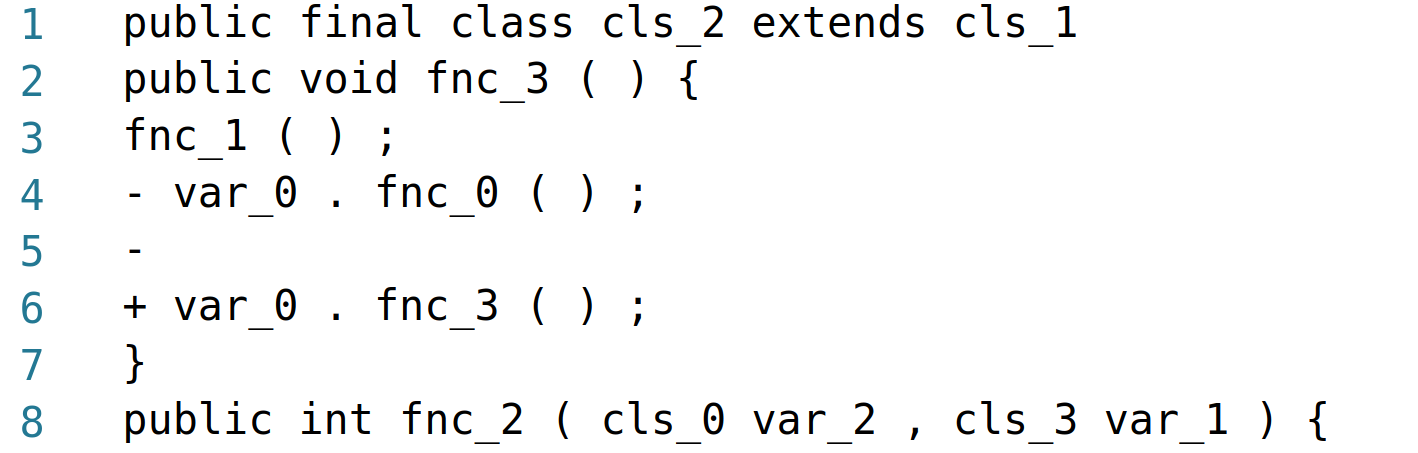}
  \end{center}
  \caption{Example of a preprocessed diff using the encoding sketch}
  \label{fig:pre-sketch}
\end{figure}

The sketch-based solution consists of three parts. First, the dataset is preprocessed. 
Constants, variables, functions, and class names are replaced with placeholders. Then the 
new dataset with placeholders and smaller vocabularies is trained. Lastly, the obtained 
messages are postprocessed and the placeholders in messages are replaced with constants, 
variables, functions, and class names extracted from diffs. An example of a prepossessed diff
can be found in Figure \ref{fig:pre-sketch}.

The preprocessing part relies on Java naming conventions in order to distinguish between 
constants, variables, functions, and class names. First, the Java keywords,
annotations, imports, numbers, strings and comments from the diff file are removed. Afterwards the following heuristics are used to distinguish between constants, variables, functions, and class
names:  

\begin{enumerate}
  \item constants are written using only upper case letters

  \item classes start with an uppercase letter and continue with a combination of upper 
  and lower case letters

  \item functions start with a lower case and have an open bracket at the end of the name
  
  \item variables start with a lower case but have no open bracket at the end of the name

\end{enumerate}
In the end, each constant, variable, function, and class name found in the diff file are
replaced with a specific placeholder. If any name found in the diff file is found in the
message, that name is also replaced with same placeholder as in the diff file. A dictionary is
created in order to save the mapping from placeholders to the actual names.  

Our sketch preprocesing part is based on the Java naming convention is reasonable but risky. A more reliable solution would have been using an AST tree to find the constants, variables, functions, and class names. Obtaining the AST from a diff file can be challenging or maybe impossible that is why we recommend using an AST tree in the future research but in our research we used our rule-based preprocessing.

The postprocessing part searches for placeholders in the predicted messages and replaces
those placeholders with the mapping found in the dictionary created during preprocessing.
If no mapping is found, a random name is selected from the names found in that diff. In 
case no name is found, the placeholder is removed.

This method has two main advantages. First of all, it reduces the vocabulary size by replacing constant, variable, function, and class names with placeholders. Secondly, it enhances the similarity between parts of code which otherwise would be very different because of the constant, variable, function, and class names.

\section{Experimental setup}\label{ES}

In this section we briefly present the datasets, training setup, and evaluation setup used during our experiments.

\subsection{Datasets}
Since the previous work on this topic was done on the dataset introduced by Jiang et al. 
\cite{Towards}, we are going to conduct most of our experiments on the same data. In order to 
be able to present a broader comparison with the state of the art we are going to use the cleaned version of Jiang et al. dataset introduced by Liu et al. \cite{How}. In the next
paragraphs we are going to present a brief description of both datasets.

\begin{figure}[htbp]
  \begin{center}
  \includegraphics[width=\linewidth]{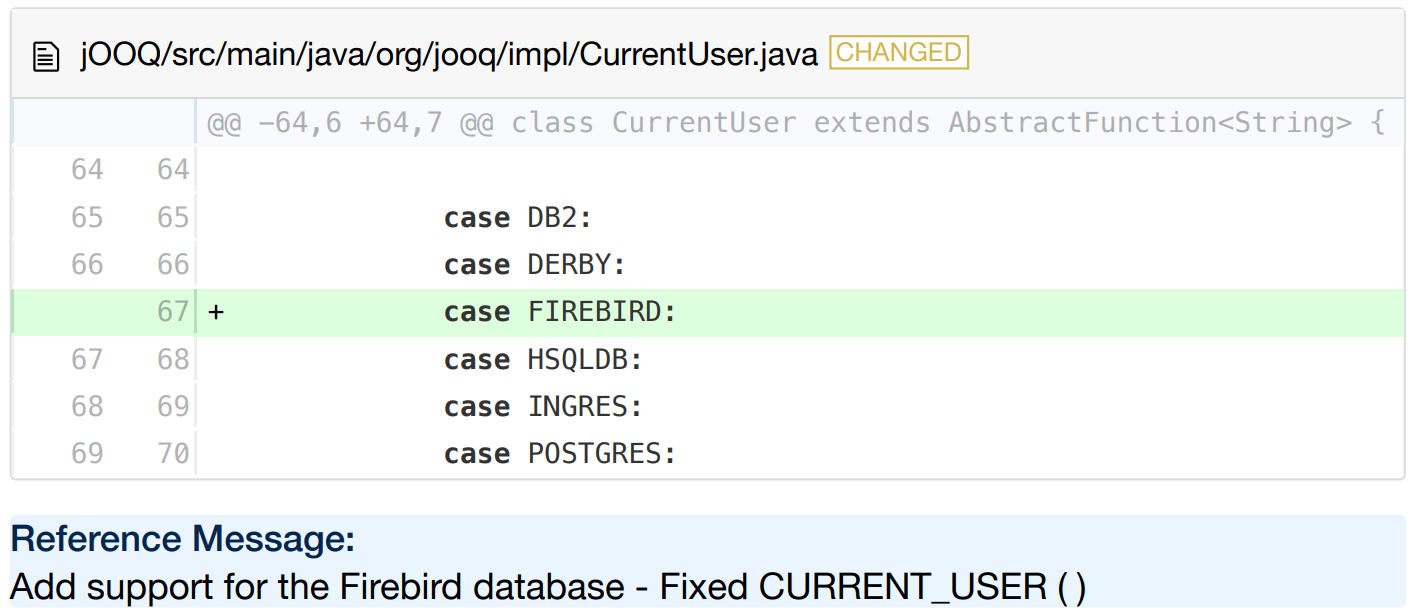}
  \end{center}
  \caption{Example of a diff and reference message from \cite{How}}
  \label{fig:automatically-d}
\end{figure}

\paragraph{Jiang el al. dataset \cite{Towards}} The dataset was created by collecting 2 
million commits from 1,000 Java repositories. The full list of projects used in creating the dataset can be found online \cite{JiangWeb}.
The commits that were rollbacks, merges or have empty messages or non-English
characters were removed. The diff files were obtained using \textit{git diff} command, but commits
containing diff files that were larger than 1 MB are filtered. In order to obtain messages
with  a consistent writing style, only messages that start with \textit{dobj}  (direct object) dependency were selected
using Stanford CoreNLP\cite{CoreNLP} framework. From the remaining
number of commits, 26,000 are randomly selected for training, 
3,000 are selected for test, and 3,000 for validation. Figure \ref{fig:automatically-d} represents an example of a diff and a reference message.

\paragraph{Liu et al. dataset\cite{How}} is a subset of the original Jiang et al. \cite{Towards} dataset. 
In this paper, we refer to this new dataset as the \textit{cleaned} dataset as it removes simple examples from the \textit{original} set. Liu et al. analyzed  the results
obtained by Jiang et al. \cite{Automatically}. They discovered that most of the high quality messages
obtained using Jiang et al. approach have high similarity with messages from the 
training set and are not describing the actual changes (e.g. they are "housekeeping" messages with trivial changes to non source-code / Java files). 
After doing an exhaustive search they identified a couple of trivial message
patterns and removed all messages which were fitting those patterns. The full list of patterns can be
found online \cite{LiuWeb}. However, as they acknowledged in their paper, this
approach did not clean all the trivial commit messages. This dataset contains 22,112 examples in the training set, 2,511 examples in the validation set, and  2,521 examples in testing set.

\subsection{Training Setup}
The training goal of our system is to minimize the categorical 
cross-entropy. As training algorithm we use Adam with a learning
rate of 0.0001. We used sequence-to-sequence models in our work with
RNN encoder and decoder. The encoder and the decoder use GRU \cite{GRU} units. 
Our models use Bahdanau attention\cite{Attention} for the attention mechanism. 
The size of the hidden units is 1024 and the size of the batches 
is 32. The models are trained between 5,000 - 30,000 steps 
depending on the number of layers of each specific model and the size of the dataset. 
We use the same input sequence length (100) and output sequence 
length (30) as Jiang et al. \cite{Towards} 
as 99\% of the examples in the datasets have both the diffs and commit messages length less than these thresholds.

The training was performed on two types of GPUs: 12GB GeForce GTX TITAN X and 12GB Tesla K40m.

\subsection{Evaluation Setup}
We evaluate the models based on the similarity between the target
messages and the generated messages. In order to evaluate the similarity
between two sentences we compute BLEU scores \cite{BLEU}.
BLEU is widely used as a similarity metric in machine translation 
\cite{Towards, Massive, Google, Attention, Copying} and various other language and code generation tasks \cite{sketch2}.
BLEU score is a number between 0 and 100 and is calculated based on n-gram overlap. 
We decode our examples using beam search, with widths of 1, 5, 10, and 10 with length penalty of 1.
Beam search is commonly used to maximize the scoring function for
target sequences in neural decoders using RNNs. 

All our experiments were run five times and the average BLEU-4 score over all runs is
presented in the results section.

\section{Experiments and Discussion}

\subsection{Baseline NMT Method} 
The experiments in this section were done on the initial dataset 
introduced by Jiang et al. \cite{Towards}.

\subsubsection{Architecture exploration}

As we stated before, we explore the following three verticals that may increase the performance of our baseline: model depth, embedding dimensionality, and residual connections.

\paragraph{Model Depth}
We compare the results obtained with
NMT models using 2, 4, and 8 layers.
Our results, presented in Table \ref{experiments:model-depth-explorarion}, 
show that a larger model is able to encode more
information and to make better predictions. 
The problem of generating commit messages out 
of diffs is difficult enough and the dataset contains sufficient examples so the larger models do not overfit. 
All the models obtain the best results using beam 
decoding of width 10 and length penalty 1.
The model with 8 layers (4 layers for encoder, 4 for decoder) obtains the best BLEU-4 score (34.07).

\begin{table}[htbp]
\caption{Influence of model depth for NMT models}
\begin{center}
\begin{tabular}{|c c| c |}
\hline
Encoder & Decoder & BLEU-4 \\
\hline
1 & 1 & 31.16 \\
2 & 2 & 33.56 \\
4 & 4 & \textbf{34.07} \\
\hline
\end{tabular}
\end{center}
\label{experiments:model-depth-explorarion}
\end{table}

\paragraph{Embedding Dimensionality}

We compare the results 
obtained with models with embedding size 1024 
and models with embedding size 512. We expected 
that a larger embedding is able to capture a better mapping 
between the words and the vector space. The 
models with embedding 512 have better performances than the others
when the decoding is done without beam search. However, when the
decoding is done with beam search the models with embedding 1024
have better performances in most of the cases. In our opinion, the suspect that the model with 8 layers and 1024 embedding size was not fully trained at
30,000 steps and can achieve a higher score than the model with 512 embedding.
In general, models with 1024 embedding need more training 
steps than the models with 512 embedding. 
We also did not observe overfitting with a larger embedding space.
In conclusion, a 1024 embedding size is more suitable for converting
diffs into commit messages as can be seen in Table \ref{experiments:embedding-dimensionality-explorarion}.

\begin{table}[htbp]
\caption{Influence of embedding size for NMT models}
\begin{center}
\begin{tabular}{|c c c| c c |}
\hline
&&&\multicolumn{2}{c|}{BLEU-4}\\
Encoder & Decoder & Embedding size &  wo BS & w BS \\
\hline
1 & 1 & 512 & 29.98 & 30.05 \\
1 & 1 & 1024 & 27.11 & \underline{31.16} \\
2 & 2 & 512 & 33.16 & 33.29 \\
2 & 2 & 1024 & 32.77 & \underline{33.56} \\
4 & 4 & 512 & 33.41 & \underline{34.17} \\
4 & 4 & 1024 & 32.58 & 34.07 \\
\hline
\end{tabular}
\end{center}
\label{experiments:embedding-dimensionality-explorarion}
\end{table}

\paragraph{Residual Connections}

We compare the results 
obtained with NMT models with residual connections 
and vanilla NMT models. 
Residual connections help the gradients flow faster through the model
and we expected these models to have better performance. A better 
gradient flow thorugh a neural model helps them to converge faster. Our intuition
is confirmed by the results. The models with residual connections
have on average an increase of 3 points in BLEU-4 score compared to the models
without residual connections (see Table \ref{tab:residual-explorarion}). The biggest impact of 
residual connections is seen for the larger NMT models, with 8 layers, which is an increase of 4.52 in BLEU-4 score.

\begin{table}[htbp]
\caption{Influence of residual connections for NMT models}
\begin{center}
\begin{tabular}{|c c c| c |}
\hline
Encoder & Decoder & Residual & BLEU-4 \\
\hline
2 & 2 & No & 33.56 \\
2 & 2 & Yes & \underline{35.35} \\
4 & 4 & No & 34.07 \\
4 & 4 & Yes & \underline{38.45} \\
\hline
\end{tabular}
\end{center}

\label{tab:residual-explorarion}
\end{table}

In the next experiments we are going to use the following three models with an embedding size of 1024: 1) \textbf{NMT2} which is a two layer model 2) \textbf{NMT4} which is a 4 layer model with residual connections, and 3) \textbf{NMT8} which is a 8 layer model with residual connections. The results obtained by the three models are presented in Table \ref{tab:models-exploration} and the last model can be considered as a strong NMT baseline for commit message generation.

\begin{table}[htbp]
\caption{Overview of three proposed NMT models}
\begin{center}
\begin{tabular}{|l | c | c |}
\hline
Model & Type & BLEU-4 \\
\hline
NMT2 & 2 layers, no residual & 31.16 \\
NMT4 & 4 layers, residual & 35.35 \\
NMT8 & 8 layers, residual & 38.45 \\
\hline
\end{tabular}
\end{center}

\label{tab:models-exploration}
\end{table}

\subsubsection{Rare and OOV Words}
\paragraph{Byte Pair Encoding (BPE)}

BPE preprocessing has a big impact on language translation. 
However, for our problem BPE achieves poor performances and obtains a BLEU-4 score of 32.67 (vs. 38.45).
The model trained with the biggest vocabulary size has the best 
performances. Our BPE vocabulary contains a lot of words which 
are rare. Because of the rarity of the words this preprocessing 
method has poor performance on this dataset. BPE shows better
performances on a dataset where rare words are not so frequent.
In the Jiang et al \cite{Towards} dataset, out of 66,295 words in the joint
target and source vocabulary, 29,941 appear only once which is 45\%
of the total of words in the joint vocabulary. The results are
presented in Table \ref{tab:bpe-results}.

\begin{table}[htbp]
\caption{Influence of BPE for NMT models}
\begin{center}
\begin{tabular}{|l | c c  c |}
\hline
 & Model & Vocabulary & BLEU-4 \\
\hline
BPE1 & NMT4 & 5,000 & 26.65 \\
BPE2 & NMT8  & 10,000 & 29.40 \\
BPE3 & NMT8 & 32,000 & \textbf{32.67} \\

\hline
\end{tabular}
\end{center}
\label{tab:bpe-results}
\end{table}

\paragraph{Copying mechanism}

As already mentioned, the dataset does not contain any
OOV words. However, we observed that 45\% of
the words in the vocabulary appear only once. This encouraged us
to reduce the vocabulary size and employ a copying mechanism for words
which became out of vocabulary after vocabulary reduction. We tested our two configurations
of vocabulary reduction on the three proposed models (NMT2,
NMT4, NMT8) and we present the results in Table 
\ref{tab:copying-results}. Vocabulary reduction and copying 
mechanism help models with fewer layers to achieve scores closer 
to those obtained with more complex neural architectures. 
Models trained using the second configuration, where
we reduce both the messages and diffs vocabularies, have performed better than
models trained using the first one. As we stated in the Proposed method section \ref{sec:proposed-method},
the first configuration employs a modified messages vocabulary which contains words that appear at least once and 
the second configuration employs vocabularies contain words that appear at least ten times.
This happens mainly for two reasons: 1) it reduces
the noise in the input sequence and the embedding layer can capture 
a better mapping 2) reduces the number of parameters for the models. 
The NMT2 and NMT4 models with vocabulary reduction and copying 
obtain better performances than the NMT2 and NMT4 models without.
However, NMT8 without vocabulary reduction and copying mechanism
obtains a better performance. We assume that NMT8 is 
able to extract some information from the noise due to the fact of having more layers. 
In most of the cases, beam search with width 10 and penalty 1 with 
copying mechanism is the decoding algorithm that achieves the best result.
The impact of the copying mechanism is limited but it increases the 
BLEU score in average by 0.22 in BLEU-4 score. 

\begin{table}[htbp]
\caption{Copying mechanism and vocabulary reduction for NMT models}
\begin{center}
\begin{tabular}{|l c c | c |}
\hline
Model & Configuration & BLEU-4 \\
\hline
NMT2 & 1 & 33.01 \\
NMT4 & 1 & 36.25 \\
NMT8 & 1 & 36.87 \\
NMT2 & 2 & 33.58 \\
NMT4 & 2 & 37.36 \\
NMT8 & 2 & \textbf{37.39} \\
\hline
\end{tabular}
\end{center}
\label{tab:copying-results}
\end{table}

In this subsection we investigated RQ1 and we discovered a more suitable NMT architecture which achieves a BLEU-4 score close to the current state of the art (NMT8, residual, no copy, no BPE).

\subsection{Improved Ensemble NMT Model}

In this section we present the results obtained by our improved ensemble method. Our experimental results are presented in Table \ref{tab:file-type} and were performed on the original dataset. NMT2 is a two layer RNN architecture trained on the entire dataset. NMT2-FT, NMT4-FT, NMT8-FT are the 2, 4, and 8 layer RNN architectures trained on datasets which contain only examples specific to a specific file type (FT). 

We compare the results obtained by training on the nine file type datasets (Gitignore, Gitrepo, Gradle, Java, Md, Properties, Txt, Xml, Yml) with the results obtained with the model with two RNN layers (NMT2) on the entire dataset. 

We firstly observe that for Gitrepo files, both the model trained on the entire dataset and the one
trained only on this specific file type obtain around 99 BLEU-4 score. This means that both approaches 
were able to solve almost perfectly examples for Gitrepo files. The very high BLEU-4 score on this file type is an outlier compared with the scores obtained on the other file types. This score is artificially increasing  the average BLEU-4 score on the entire dataset and it contains only trivial, "maintenance" commit messages.

On the datasets containing Gradle, Java, Md, and Xml file types, the performance of NMT2-FT is on average 161\% higher than the performance of the baseline NMT2. The increase in performance of Gitignore, Properties, Txt is a bit lower (117\%). This is mostly because of the small number of examples for these datasets. On the dataset containing Yml examples NMT2 is not able to learn, but NMT2-FT  is able to achieve a BLEU-4 score of 6.42.

We observe that deeper models (NMT4-FT and NMT8-FT) have better performance. However, the increase of performance is limited compared to the increase of performance in the case of training on the entire dataset. On average a four layer model increases the performance with 2.5\% and a eight layer model with 6.9\%.

In comparison with NNGen, NMT2-FT has similar performances on examples containing Gitrepo and Gradle files, a slightly better one on Gitignore, Md, Properties, Xml files and a slightly worse on Yml files. However, NNGen is still outperforming our ensemble approach on Java and Txt files by a high margin (13.1 vs. 9.06, 22.57 vs. 8.11).

\begin{table}[htbp]
\setlength\tabcolsep{4.5pt}
\caption{Ensemble NMT model results by file type}
\begin{center}
\begin{tabular}{|l | c c c c c |}
\hline
  & NNGen\cite{How} & NMT2 & NMT2-FT & NMT4-FT & NMT8-FT  \\
\hline
Gitignore & 5.47 & 2.48  & \underline{6.2} & - & -  \\
Gitrepo & \underline{100} & 99.64 & 99.21 & - & -  \\
Gradle & 25.11 & 11.29 & 25.15 & 25.51 & \underline{26.76} \\
Java & \underline{13.1} & 2.04 & 9.06 & 9.38 & 9.70 \\
Md & 18.96 & 11.52 & \underline{21.46} & - & - \\
Properties & 19.26 & 9.05 & \underline{21.33} & - & - \\ 
Txt & \underline{22.57} & 8.11 & 13.69 & - & - \\
Xml & 22.43 & 7.14 & 26.46 & 27.18 & \underline{28.4} \\
Yml & \underline{8.69} & 0 & 6.85 & - & - \\
\hline
\end{tabular}
\end{center}
\label{tab:file-type}
\end{table}

In Table \ref{tab:assemble-file-type} we present the results obtained with the assembles of NMT2-FT,
NMT4-FT and NMT8-7 in both scenarios in comparison with NNGen, NMT2 and NMT8. NMT4-FT-S1 and NMT4-FT-S2 are
ensembles containing NMT models with four layers trained on the Gradle, Java, Xml and Others datasets and models with two layers for the rest of the datasets. Similarly, NMT8-FT-S1 and NMT8-FT-S2 contain eight layers models trained on the Gradle, Java, Xml and Others datasets and models with two layers for the rest of the datasets. The description of the scenarios can be found in the following subsection \ref{subsec:sketc-proprosed}.

The ensemble of NMT2-FT models is slightly outperformed by NNGen on both scenarios (38.55 vs. 38.33, 38.44). Ensembles of NMT4-FT and NMT8-FT models were able to outperform NNGen due to the performance increase on Gradle, Java and Xml datasets. However, on a more thorough analysis NNGen still outperforms our approach for Java source files. 

The performance of the ensemble containing all the NMT2-FT models in the first scenarios (NMT2-FT-S1) is with 7.17 greater in BLEU-4 score than the performance of the NMT2 and with 7.28 in the second scenario (NMT2-FT-S2). The score obtained with NMT8 (8 layer model) is a bit better than the one obtained with NMT2-FT-S1 (38.44 vs. 38.33) and similar to the one obtained with NMT2-FT-S2 (38.44). We believe that NMT8 is able to separate the word distributions by file type due to its larger number of layers.

In both scenarios, the ensemble containing NMT8-FT is achieving a BLEU-4 score of 39.01. It is
fair to say that training Gitignore, Properties, Txt and Yml separately does not boost the performance. This is mainly due to the reduced size of those datasets. However, if the dataset size is large enough we would recommend training examples containing diffs with similar file type separately. 

\begin{table}[htbp]
\caption{Ensemble models compared to state of the art}
\begin{center}
\begin{tabular}{|l c|}
\hline
Approach & BLEU-4  \\
\hline
NNGen\cite{How} & 38.55 \\
NMT2 & 31.16 \\
NMT8 & 38.44 \\
NMT2-FT-S1 & 38.33  \\
NMT2-FT-S2 & 38.44  \\
NMT4-FT-S1 & 38.65 \\
NMT4-FT-S2 & 38.69  \\ 
NMT8-FT-S1 & \textbf{39.01}  \\
NMT8-FT-S2 & \textbf{39.01}  \\
\hline
\end{tabular}
\end{center}
\label{tab:assemble-file-type}
\end{table}

The experiments in this section revealed that training on a specific file type improves overall performance (RQ2). Most of the ensembles outperformed the state of the art on the original dataset. 

\subsection{Sketch-based Neural Encoder}\label{subsec:sketch}

We trained our sketch-based method on three different architectures. Table \ref{tab:Java-results} 
presents the results in BLEU-4 score for NNGen, NMT8-FT, and our sketch-based
method using models with 2 (NMT2-JT), 4 (NMT4-JT) and 8 (NMT8-JT) layers. The sketch-based method increases the performance of NMT models by 40\% for Java files. It also reduces 
the complexity of the problem and models with fewer layers have similar or even better
performances than deeper models. It is interesting that the model with 2 layers can achieve
a better BLEU-4 score than the model with 4 layers. However, in average across all prediction
strategies (without beam, beam 5, beam 10, beam 10 with length penalty 1) NMT4-JT archives 
a score of 13.38 and NMT2-JT achieves a score of 13.32. 
The score obtained with NMT8-JT is better with 4\% than the score in BLEU-4 obtained with 
NNGen on diffs containing Java files. 

\begin{table}[htbp]
\caption{Sketch-based NMT model for Java files}

\begin{center}
\begin{tabular}{| l c |}
\hline
Approach & BLEU-4 \\
\hline
 NNGen\cite{How} & 13.1 \\
 NMT8-FT & 9.7 \\
 NMT2-JT & 13.6  \\
 NMT4-JT & 13.47  \\
 NMT8-JT & \textbf{13.64}  \\
\hline
\end{tabular}
\end{center}
\label{tab:Java-results}
\end{table}

In this section we showed the positive impact of the sketch-based encoder for NMT models (RQ3). We improved our score by a high margin and outperformed the state of the art on Java files. At this moment, NMT sketch-based models achieve better results both overall, for all file types in the dataset, and for Java files alone.  

\section{Comparison with state of the art}

In this section we compare the results obtained with our eight layer model trained on the entire dataset (NMT8), our ensemble of models trained on specific file types (NMT8-FT) and the same ensemble but with Java sketch-based approach (NMT8-FT-JT) with the current state of the art.

The first, third, forth and fifth row in Table \ref{tab:comparison} show how our improved approaches are
outperforming by a high margin the state of the art in generating commit messages from
diffs using vanilla NMT techniques\cite{Automatically} on both the initial and the cleaned dataset. 

Our file type (NMT8-FT) and Java sketch (NMT8-FT-JT) approaches have better results that the NNGen approach introduced by Liu el al. \cite{How} and even our baseline model with 8 layers (NMT8) has similar results with NNGen on
the original dataset. However, on the cleaned one only our Java sketch (NMT8-FT-JT) is outperforming NNGen.

Although NNGen has good results on these two datasets, we believe the nearest neighbour approach
will lack robustness on larger and more difficult datasets.  We also highlight the fact that the reduced number of examples in our
file type datasets has a negative impact on the performance of our NMT approaches. Thus, the performance gap between our proposed NMT sketch-based ensemble approach and NNGen will probably increase on larger and more complex datasets.

Another important aspect when we compare our results with the state of the art is the run-time performance of commit message generation. In comparison with Liu et al.\cite{How} which take into account the training time and the prediction time during during speed evaluation, we compare the average prediction time on the test dataset because it is the most relevant metric for production run-time. The results can be found in Table \ref{tab:comparison}. Our approaches are 27-35x slower than the NNGen \cite{How} but are still fast enough to be used in production. Our file type (NMT8-FT) approach is faster than our eight layer model trained on the entire dataset (NMT8) with about 50ms. This speed boost is mainly due to the vocabulary reduction on each file type. Our Java sketch (NMT8-FT-JT) is bit slower than our file type (NMT8-FT) but the boost in performance is significant. 

\begin{table}[htbp]
\caption{Comparison with state of the art}\label{SA}

\begin{center}
\begin{tabular}{| l  l c l |}
\hline
Dataset & Approach & BLEU-4 & Avg Time \\
\hline
Original & NMT\cite{Automatically} & 31.92 & 340ms\cite{How}\\
 & NNGen\cite{How} & 38.55 & \textbf{10ms}\cite{How} \\
 & Our NMT8 & 38.44 & 323ms \\
 & Our NMT8-FT & 39.01 & 256ms \\
 & Our NMT8-FT-JT & \textbf{39.54} & 271ms\\
\hline
Cleaned & NMT\cite{Automatically} & 14.19 & 309ms\cite{How} \\
 & NNGen\cite{How} & 16.42 & \textbf{9ms}\cite{How} \\
 & Our NMT8 & 16.07 & 314ms \\
 & Our NMT8-FT & 16.07 & 264ms \\
 & Our NMT8-FT-JT & \textbf{16.86} & 281ms \\

\hline
\end{tabular}
\end{center}
\label{tab:comparison}
\end{table} 

\section{Human evaluation}\label{HE}

In this section we present the human evaluation we conducted for assessing the quality of the generated commit messages. The goals of this section are to offer a more in-depth evaluation of our results and to confirm that our solution provides a real improvement.

\begin{figure}[htbp]
  \begin{center}
  \includegraphics[width=\linewidth]{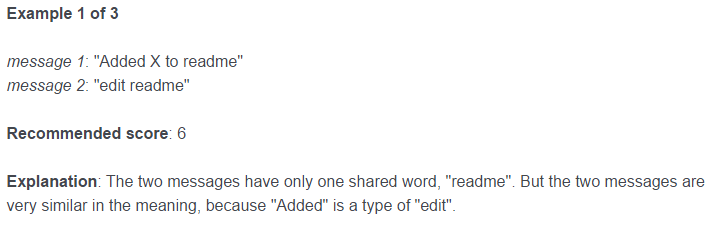}
  \end{center}
  \caption{A scoring example introduced by Jiang et al.\cite{Automatically}}
  \label{fig:scoring-example}
\end{figure}

We evaluate the results obtained by the sketch-based NMT model described in section \ref{subsec:sketch} and which obtains the best quantitative BLEU-4 scores on both datasets. Our evaluation is conducted on all 436 test examples which contain diffs of Java files. We believe and strongly encourage that future evaluations should be done explicitly on this file type. We use a similar survey design as Jiang et al. \cite{Automatically} with a score range between 0-7 where 0 means no similarity and 7 means identical. The commit messages were evaluated by 18 professionals, all of them have majored in computer science. Each of them evaluated between 36 to 50 messages according to the three examples introduced by Jiang et al. \cite{Automatically}. An example is presented in Figure \ref{fig:scoring-example}.

We group the mean scores into three groups: 1) low quality (0 - 2.33] 2) medium quality (2.33 - 4.66] 3) high quality (4.66 - 7]. Unsurprisingly, 55.73 \% of the generated commit messages were low quality, 26.16 \% medium quality, and 18.11 \% high quality. Cohen's Kappa coefficient between raters was 0.346 which suggests there was a fair agreement between them. This fair inter-rater agreement score also shows that assessing the quality of commit messages is difficult for human professionals as well.

However, the average score obtained by our generated messages is 2.40 on a 0-7 scale which is 1.50 on a 0-4 scale. This average score is slightly higher than the score obtained by Liu et al. \cite{How} on the cleaned dataset (1.46). Taking into consideration that Java file type is one of the most complex file types in the cleaned dataset, we are confident to say, according to our human evaluation, that our approach outperforms his approach on examples containing diffs of Java files and on the entire dataset. This analysis should take into account that the raters are different in the two studies. However, our study uses a large number of persons (18 professionals, graduates in computer science) to eliminate potential outlier scores.

In figure \ref{fig:comparison-example} we present three examples of generated messages. The first one was evaluated as a good quality message and the next two were evaluated as low quality messages. In the first example, the NMT model was able to predict a message similar with the target message but with a grammatical error. In the second example, the model was able to partially understand the changes. The generated message was categorized as low quality because the evaluators did not have access to the diff files during evaluation. In the last example, we present a situation where our model failed to generate a meaningful message. However, it is clear that our sketch method outperforms NNGen on the first two examples and has a similar performance on the third one.

\definecolor{my_bleu}{rgb}{ 0.66, 0.78, 1}
\definecolor{my_red}{rgb}{ 1, 0.71, 0.70}
\definecolor{my_yellow}{rgb}{ 1, 0.99, 0.66}
\definecolor{my_green}{rgb}{ 0.71, 1, 0.66}
  
\begin{figure}[htbp]
  \small{Example 1:}
  \begin{center}
  \includegraphics[width=\linewidth]{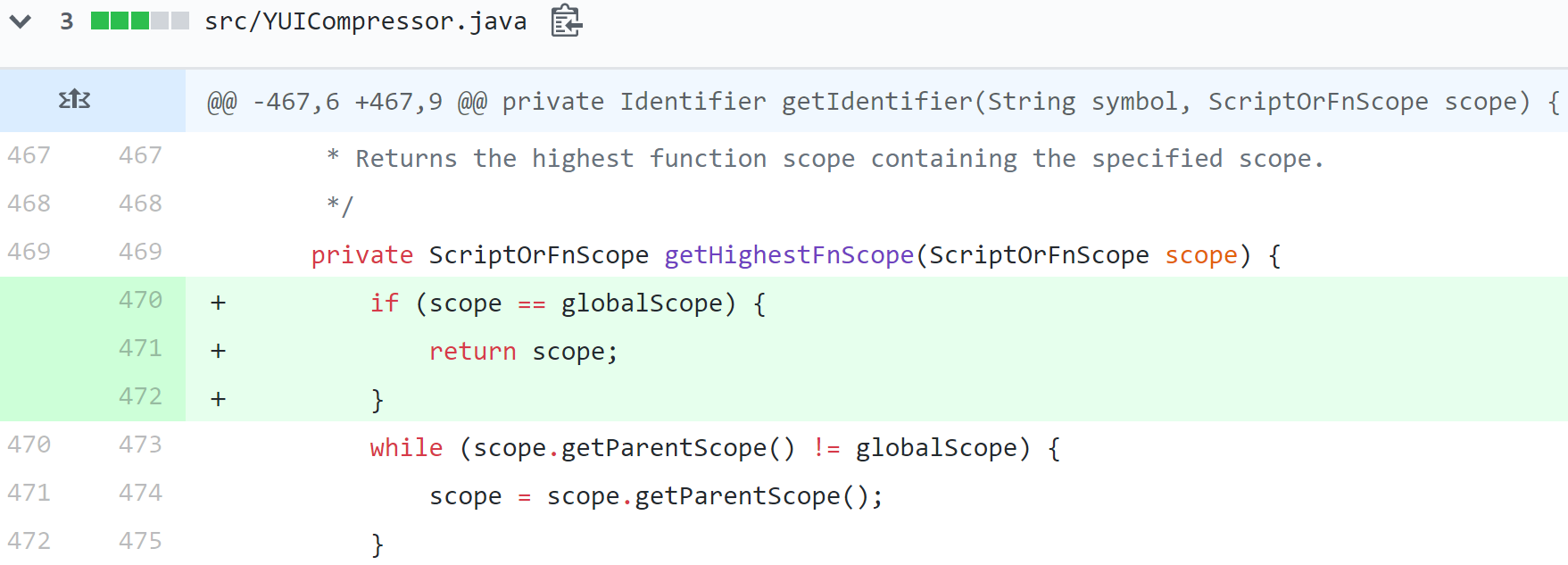}
  \end{center}
  \colorbox{my_bleu}{\footnotesize{Reference Message: Fixed problem with getHighestFnScope method}} \\
  \colorbox{my_yellow}{\footnotesize{Message Generated by NNGen: Remove fflush from hdr\_percentile\_print.}} \\
  \colorbox{my_green}{\footnotesize{Message Generated by NMT8-JT: Fix a bug with getHighestFnScope().}} \\
  \vspace{1mm}\\
  \small{Example 2:}
  \begin{center}
  \includegraphics[width=\linewidth]{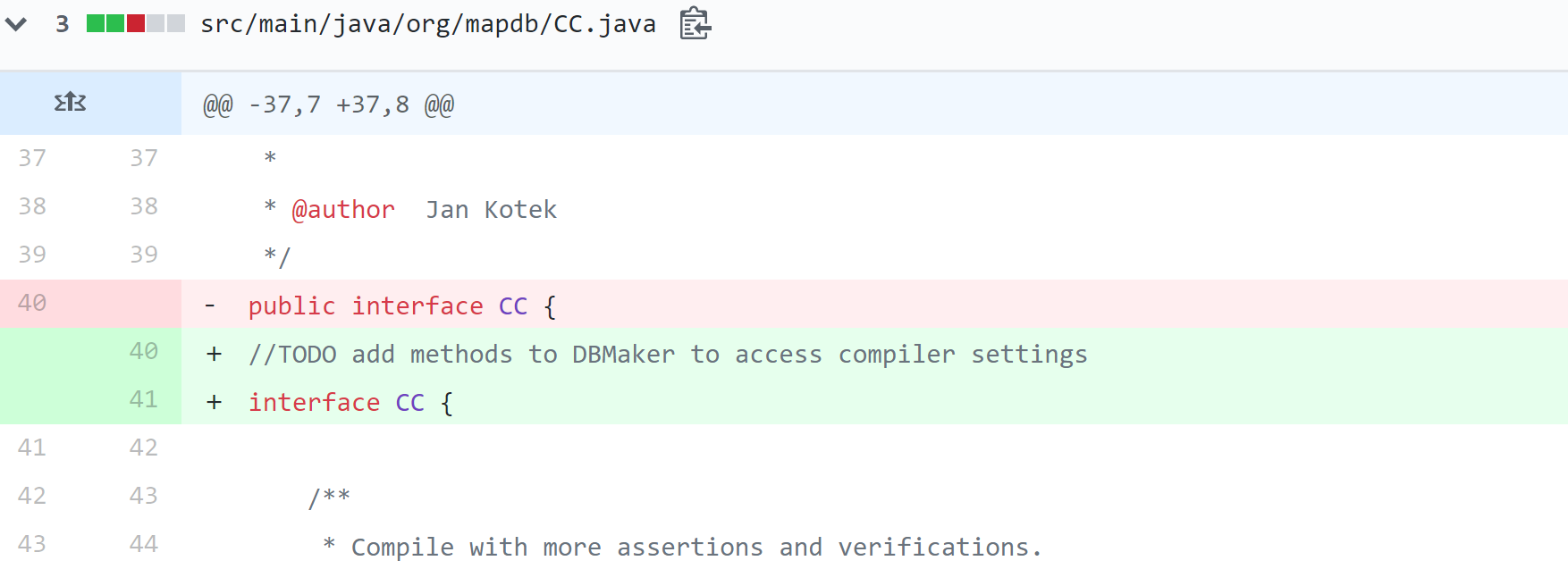}
  \end{center}
  \colorbox{my_bleu}{\footnotesize{Reference Message: make package protected, so other apps do not link it}} \\
  \colorbox{my_yellow}{\footnotesize{Message Generated by NNGen: add function with zero arguments}} \\
  \colorbox{my_green}{\footnotesize{Message Generated by NMT8-JT: add comments}} \\
  \vspace{1mm}\\
  \small{Example 3:}
  \begin{center}
  \includegraphics[width=\linewidth]{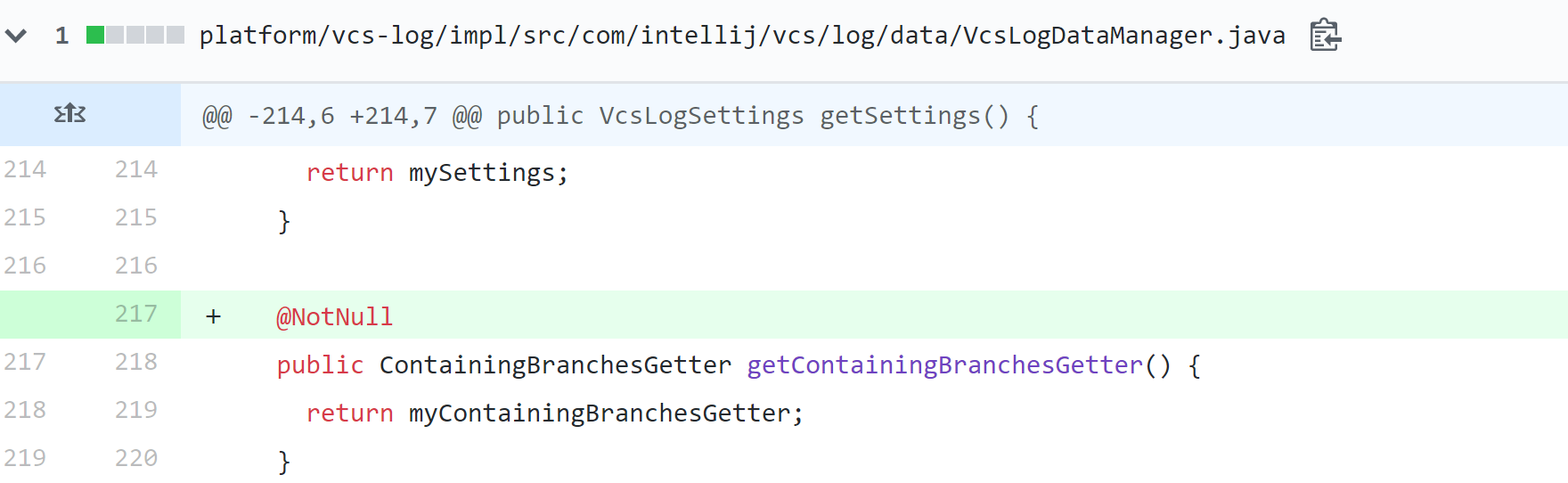}
  \end{center}
  \colorbox{my_bleu}{\footnotesize{Reference Message: add not null}} \\
  \colorbox{my_yellow}{\footnotesize{Message Generated by NNGen: set visible pack to empty in vcs log ui in}} \\
  \colorbox{my_yellow}{\footnotesize{dispose IDEA - 120539}} \\
  \colorbox{my_green}{\footnotesize{Message Generated by NMT8-JT: JMS now also enlists topic destinations}} \\
  \colorbox{my_green}{\footnotesize{as managed endpoints in JMX}}
  \caption{Examples of generated and the reference messages}
  \label{fig:comparison-example}
\end{figure}

\section{Conclusions}\label{CO}

During our work we were able to show how NMT techniques are able to improve 
the generation of comments for code commits and can achieve state of the art results. 
Our sketch-based ensemble model is outperforming all existing solutions on both datasets used to assess the quality of generated commit messages for this task.

One of our main contributions is the approach of splitting the dataset in
smaller datasets by file type and training them separately in an ensemble of NMT models. We are confident that an
increase of the number of the examples for each specific file type dataset would have 
a positive impact in performance of the overall system.

Inspired by the NMT's poor performance on Java files, we also propose a sketch-based method which is able to boost the performance on Java types by a high margin and outperform NNGen - the top scoring solution for Java files at this moment.

We believe that the problem of generating messages for code commits  is more challenging and of a somehow different nature than the problem of translating text from one natural language to another. However, the current datasets for commit message generation are 100 times smaller than the datasets used in machine translation (e.g. WMT'14 English-to-French, WMT'14 English-to-German). A larger dataset is a must in order to ensure progress on this task. We believe that NMT models did not achieve their full potential on this task and will perform even better on a larger dataset.

Diff files contain some context of the changes, but that context contains limited information. A future improvement might likely target adding more context to training process. One of the challenges of this approach will be choosing the right size for the context. At the end, we view our work as a step forward for advancing the state of the art, but current 
approaches are far from being able to be used in the industry.
\bibliographystyle{IEEEtran}
\bibliography{mybib}

\end{document}